\RequirePackage{fix-cm}
\documentclass[10pt,twocolumn,letterpaper]{article}

\usepackage[review]{cvpr}
\toggletrue{cvprfinal}
\ClearShipoutPicture

\usepackage[T1]{fontenc}
\makeatletter
\def\NAT@cmprs{0}
\makeatother
\usepackage{tabularx,array,colortbl,multirow,enumitem,placeins,float}
\definecolor{cvprblue}{rgb}{0.21,0.49,0.74}
\usepackage[unicode,pagebackref,breaklinks,colorlinks,allcolors=cvprblue]{hyperref}
\hypersetup{pdftitle={EvoSearch: Self-Evolving Active Perception for Wide-Area Geospatial Search}}
\definecolor{ourrow}{HTML}{EBF3FA}
\newlength{\paperTableWidth}
\makeatletter
\newcommand{\paperheaderbg}[1]{}
\newcommand{\papergroupdivider}[2]{\smash{\rlap{\hspace{-0.35pt}\color{black}%
\rule[-\dimexpr#2\ht\@arstrutbox+#2\dp\@arstrutbox-\ht\@arstrutbox-3pt\relax]{0.35pt}%
{\dimexpr#1\ht\@arstrutbox+#1\dp\@arstrutbox-6pt\relax}}}}
\newcommand{\paperheaderdivider}{\smash{\rlap{\hspace{-0.35pt}\color{black}%
\rule[\dimexpr1pt-\dp\@arstrutbox\relax]{0.35pt}%
{\dimexpr2\ht\@arstrutbox+2\dp\@arstrutbox+4pt\relax}}}}
\newcommand{\paperbodydivider}[2][2.2pt]{\smash{\rlap{\hspace{-0.35pt}\color{black}%
\rule[\dimexpr1pt-\dp\@arstrutbox\relax]{0.35pt}%
{\dimexpr#2\ht\@arstrutbox+#2\dp\@arstrutbox-#1-1pt\relax}}}}
\makeatother

\newcommand{\compactmethodgroup}[1]{\noalign{\vskip1pt%
\hbox to\paperTableWidth{\hskip\tabcolsep\normalfont\itshape #1\hfil}\vskip1pt}}

\newcommand{\methodindent}{\ignorespaces}
\newcommand{\variantindent}{\ignorespaces}

\usepackage{stfloats}

\def\confName{CVPR}
\def\confYear{2027}
\title{EvoSearch: Self-Evolving Active Perception for Wide-Area Geospatial Search}
\author{\smash{\begin{tabular}{@{}c@{}}\normalsize Zhenghao Xie\textsuperscript{1}, Jing Xiao\textsuperscript{1}, Zihao Yu\textsuperscript{1}, Wanyi Yu\textsuperscript{1}, Liang Liao\textsuperscript{2}, Gui-Song Xia\textsuperscript{1}, Mi Wang\textsuperscript{1}\\[3pt]
\normalsize \textsuperscript{1}Wuhan University\qquad\textsuperscript{2}Xidian University\end{tabular}}}
\hypersetup{pdfsubject={Preprint},pdfauthor={Zhenghao Xie, Jing Xiao, Zihao Yu, Wanyi Yu, Liang Liao, Gui-Song Xia, Mi Wang}}
\makeatletter
\g@addto@macro\@maketitle{\begin{minipage}{\textwidth}
\centering
\captionsetup{type=figure}
\includegraphics[width=\linewidth]{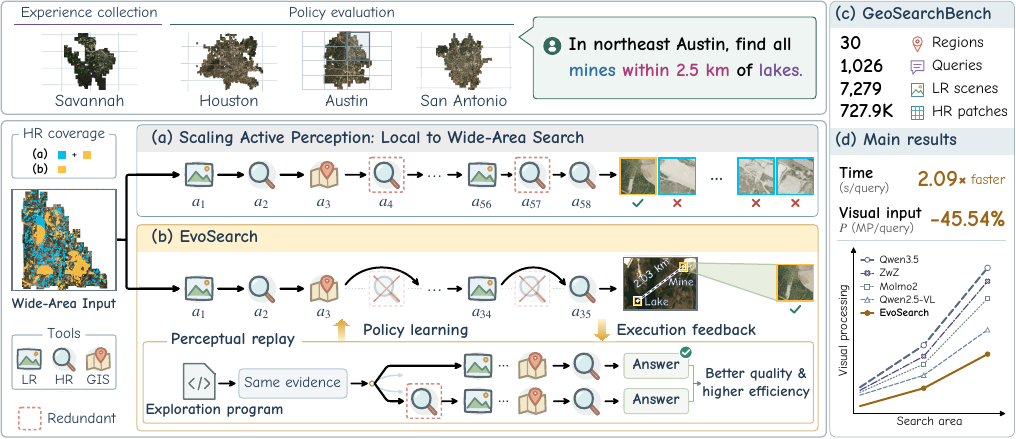}
\caption{\textbf{EvoSearch for wide-area geospatial search.} (a) Redundant observations. (b) Learning observation and stopping from branch quality and cost under shared evidence. (c) GeoSearchBench statistics. (d) Gains over initialization on GeoSearchBench, with schematic scaling curves. LR/HR denote low/high resolution, MP megapixels, and GIS a geographic information system.}\label{fig:showcase}
\end{minipage}
\par\medskip
}
\makeatother
\begin{document}
\raggedbottom
\setcounter{dbltopnumber}{1}
\setcounter{topnumber}{1}
\widowpenalty=10000
\clubpenalty=10000
\maketitle
\begin{abstract}
Geospatial search is fundamental to resource surveys and environmental monitoring, requiring the discovery of relevant evidence across large geographic areas and multiple spatial scales. Existing active perception methods adaptively select observation locations and scales to improve search efficiency, yet their search trajectories become increasingly long and costly as the search area expands. A key limitation is that final-answer correctness alone provides little guidance on which observations to retain and when to stop, hindering efficient wide-area search. To address this challenge, we propose EvoSearch, a self-evolving active perception framework that co-evolves the perception policy and exploration program to compress search trajectories while preserving answer quality. Specifically, quality--cost comparisons between completed branches from shared evidence train the policy to acquire useful observations, avoid redundant ones, and stop when sufficient evidence is available. Execution feedback from the updated policy then guides exploration-program revision, adapting experience generation to the policy's changing observation needs. We further introduce GeoSearchBench, a benchmark for evaluating search quality and visual processing efficiency through relational search and region search. Experiments demonstrate that EvoSearch shortens search trajectories and substantially reduces visual processing while maintaining overall search quality, supporting scalable active perception for wide-area geospatial search.
\end{abstract}

\section{Introduction}
Geospatial search is important for resource surveys and environmental monitoring. By finding targets that satisfy spatial constraints, it helps describe their distribution and relationships with the surrounding environment. For example, surveying industrial facilities near rivers requires finding scattered facilities throughout a study region and verifying their distances to rivers. This requires agents to acquire evidence across locations and scales while balancing search coverage and target detail. However, a low-resolution (LR) overview can obscure small targets, while exhaustive high-resolution (HR) processing is costly~\cite{xie2026observe}. Active perception allows agents based on vision-language models (VLMs) to select observation locations and resolutions on demand, providing a way to coordinate spatial coverage and local detail~\cite{r2,r3,r6}. As search areas expand, allocating observations effectively while preserving answer quality and controlling visual processing cost becomes a key challenge for scalable geospatial search.

{
To meet this challenge, agents must coordinate target discovery, spatial verification, and further search across locations using accumulated evidence. They must also determine when the evidence is sufficient to answer. However, learning these choices requires experience that distinguishes necessary from avoidable observations. Existing methods learn observation choices from demonstrations and interaction rewards~\cite{wang2026seekvln,yao2026activemopt}. Others use verbal feedback or accumulated experience to guide decisions~\cite{r29,r8,r30,r9}. However, task completion alone does not show which observations are useful. Search trajectories can reach the same correct answer at different costs, so final-answer supervision is insufficient to directly guide intermediate observation decisions. Directly imitating successful trajectories may therefore retain unnecessary observations. The resulting overhead accumulates and hinders scalability. As Fig.~\ref{fig:showcase}(a,b) illustrates, more observations do not necessarily improve final-answer accuracy.\par}

{
To address this supervision gap, we assess observations through their effects on answer quality and cost. We therefore derive decision supervision from trajectory-compression attempts. If fewer observations or earlier stopping preserve answer quality, the attempt identifies avoidable cost. If answer quality declines, it instead indicates which observations to retain; higher-budget alternatives can also identify useful additional evidence. Since omitting an observation may require later acquisitions, compression must be assessed over complete executions. Through these comparisons, successful and unsuccessful attempts inform later observation decisions.\par}

{
Building on this idea, we propose \textbf{EvoSearch} for policy--exploration co-evolution. Specifically, \emph{perceptual replay} explores stopping and visual-budget alternatives from shared evidence, executes each branch to a final answer, and compares their answer quality and total observation cost. These comparisons provide decision supervision. However, policy updates change trajectories and compression opportunities. Experience generation must therefore adapt to the current policy. Inspired by feedback-driven program improvement~\cite{r10,wu2026rsisurvey,srikanth2026aide2}, \emph{meta-exploration} uses execution feedback from the updated policy to revise and select exploration programs that seek informative comparisons. These comparisons further guide policy learning, while the updated policy supplies states and execution feedback for subsequent exploration. This alternation couples observation learning with experience generation, drawing on work that improves agents and their update procedures~\cite{r44,zhang2026hyperagents}.\par}

{
To study this problem, we further introduce a benchmark, \textbf{GeoSearchBench}, to jointly evaluate geospatial search quality and processed visual-input volume. Specifically, relational search returns target sets satisfying spatial constraints, while region search returns regions satisfying the query. Together, these tasks assess cross-location evidence acquisition, spatial verification, and stopping decisions. Experiments on GeoSearchBench show that EvoSearch reduces processed visual-input volume while maintaining overall answer quality. Furthermore, we analyze quality and cost across search scopes and use ablations with the same number of policy-training updates across the compared policy variants to examine decision supervision and exploration-program revision. Our contributions are summarized as follows:\par}
\begin{itemize}[leftmargin=1.4em,topsep=3pt,itemsep=2pt,parsep=0pt]
\item { We study how remote-sensing agents conduct wide-area geospatial search, linking multi-scale evidence acquisition with spatial reasoning to extend active perception beyond local scenes while tracking observation cost.\par}
\item We propose \textbf{EvoSearch}, a self-evolving framework that supervises observation decisions through perceptual replay and adapts experience generation through meta-exploration, coupling policy learning with feedback-guided exploration-program revision.
\item We introduce \textbf{GeoSearchBench} to evaluate geospatial search quality and processed visual-input volume through relational search and region search.
\end{itemize}

\section{Related work}
{
\noindent\textbf{Geospatial Visual Reasoning.}
Remote-sensing VLMs align imagery and language~\cite{r19,r50} for captioning, grounding, and question answering~\cite{r20,r48,r49,r23} across sensors~\cite{earthdial2025}, high-resolution imagery~\cite{r21}, and task levels~\cite{yao2026remotereasoner}. Interleaving reasoning and actions~\cite{r38} provides a general basis for tool use. In geospatial tasks, visual tools~\cite{r25}, geographic information system (GIS) operations~\cite{r26}, and geographic programs~\cite{yasuki2025geoprog3d} integrate local evidence for localization, spatial verification, and regional analysis. Meanwhile, token pruning~\cite{luo2025coarsefine,yang2025visionzip} and selective processing~\cite{hide2025,r37} reduce redundant computation, while active observation selects further evidence~\cite{r3}. Beyond these capabilities, we balance quality and cost in scalable cross-location search with spatial verification. Accordingly, GeoSearchBench assesses quality and processed visual-input volume in relational and region search.\par}

{
\noindent\textbf{Active Perception.}
Active perception acquires relevant evidence by selecting where to look~\cite{r41}, with later work also adapting scale~\cite{r39,r46}. Two main approaches are search and policy learning: language and context guide hierarchical zooming for local detail~\cite{r1,r2}; demonstrations and interaction rewards teach task-adaptive region selection and visual operations~\cite{r39,chen2025visrl,r15,r3}. Furthermore, resolution selection balances recognition quality and cost~\cite{r40,r46}, navigation locates targets sequentially~\cite{sarkar2024gomaageo,mi2025geoexplorer}, and iterative evidence acquisition alternates inspection and reasoning~\cite{shao2026rseot}. Together, these approaches advance active perception. Building on them, we learn what to observe through trajectory compression and adapt experience generation as policies change, balancing answer quality and cost.\par}

\section{EvoSearch}\label{sec:method_overview}
EvoSearch combines multi-scale observations and GIS for geospatial search, as shown in Figure~\ref{fig:framework}. During training, perceptual replay compares completed branches from shared evidence to supervise observation decisions through quality and cost differences. Meta-exploration revises the exploration program as policy updates change trajectories and exploration states.

\subsection{Formulation}\label{sec:perception_framework}
Given query $q_i$, LR imagery of search region $\Omega$, and on-demand HR imagery, geospatial search returns target or region sets satisfying the query's spatial constraints. Active perception balances answer quality and observation cost. To express this trade-off, we represent evidence acquisition and spatial operations as executable tool actions. Together, they form search trajectories subject to the query constraints, as Figure~\ref{fig:framework}(a) illustrates.

For query $i$ at decision step $t$, context $x_{i,t}=(q_i,h_{i,t},v_{i,t})$ contains the query, interaction history, and available observations. Policy $\pi_\theta$ with parameters $\theta$ selects $a_{i,t}\sim\pi_\theta(\cdot\mid x_{i,t})$. Tool feedback updates the context until answer submission. Trajectory $\tau_i$ records outcomes and observations. Let $s(\tau)$ denote answer score, $L(\tau)$ execution steps, and $H(\tau)=\sum_{t=1}^{L(\tau)}\Delta H_t$ total newly acquired HR patches, with $\Delta H_t$ counting acquisitions at step $t$. Evaluation measures processed visual-input volume $P$ from actual encoder inputs, including repeats.

We seek trajectories that preserve answer quality while reducing observation cost. Within the tools and execution budget, candidate trajectories $\Gamma(x_{i,0})$ are ranked first by quality and then by cost:
\begin{equation}
\tau_i^*\in\underset{\tau\in\Gamma(x_{i,0})}{\operatorname{arg\,min}_{\mathrm{lex}}}
\bigl(-s(\tau),C(\tau)\bigr).
\label{eq:trajectory_objective}\end{equation}
Cost $C(\tau)$ ranks trajectories with equal answer quality. Within each comparison, quality takes priority over cost. The ranking guides supervision, with quality and cost assessed jointly at evaluation. In geospatial search, $C(\tau)=(H(\tau),L(\tau))$ orders trajectories by HR acquisitions and then steps. Acquisition cost guides these preferences, while $P$ measures visual processing at evaluation. Image-based visual question answering (VQA) ranks cost by $P$.

\subsection{Scalable geospatial search}\label{sec:scalable_perception}
Wide-area search must balance spatial coverage and target detail. LR imagery may miss small targets, whereas exhaustive HR observation processes large visual inputs. Therefore, we combine both scales: LR observations provide regional overviews with less input to identify locations for closer inspection, while HR observations supply the detail needed to answer. Together, they focus detailed observations on regions relevant to the query.

\begin{figure*}[t]\centering
\includegraphics[width=\textwidth]{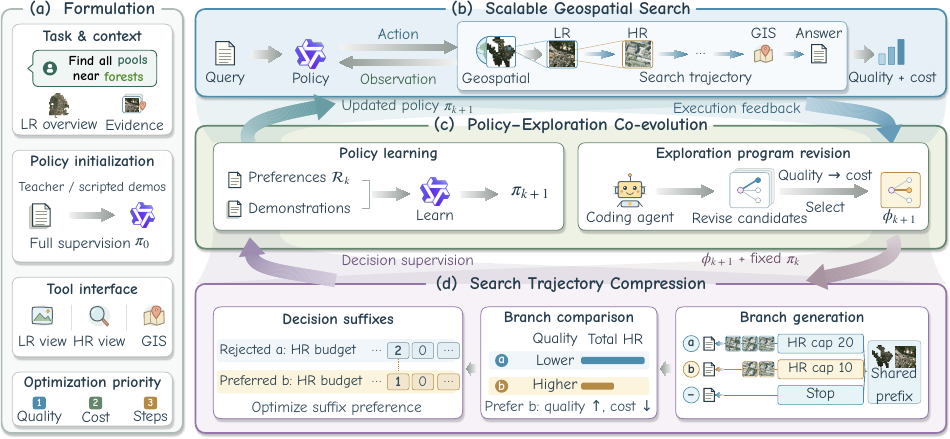}\par
\vspace{-2mm}
\caption{\textbf{Overview of the EvoSearch framework.} (a) Task formulation and initialization. (b) Geospatial search with low-/high-resolution observations and GIS. (c) Meta-exploration revises exploration programs from execution feedback. (d) Replay compares completed branches from shared evidence by quality and total HR acquisitions, then supervises decision suffixes. Here $\pi_k$ abbreviates $\pi_{\theta_k}$.}\label{fig:framework}
\vspace{-2mm}
\end{figure*}

To support this process, we organize multi-scale imagery into LR scenes and HR patches that can be acquired independently. In GeoSearchBench, the HR image covering each LR scene's geographic extent has a $10\times10$ patch grid. Each LR scene and HR patch has $268\times268$ pixels, but HR patches cover smaller areas and reveal finer details. Using their extents and cross-scale correspondence, tool interface $\mathcal T$ acquires observations on demand and connects evidence through GIS, as shown in Figure~\ref{fig:framework}(b). These fixed-size units support selective observations across locations and scales, avoiding redundant acquisitions over wider areas.

Store $W$ retains regional evidence across queries in Evidence Reuse and resets for each Single Query.

\subsection{Search trajectory compression}\label{sec:observation_comparisons}\label{sec:decision_learning}
Imitating successful trajectories may retain unnecessary observations. Perceptual replay tests which observations are needed by comparing completed branches from shared evidence by quality and cost, as shown in Figure~\ref{fig:framework}(d).

\noindent\textbf{Branch generation.} Fewer acquisitions now may require more later or leave insufficient evidence to answer. Therefore, we evaluate complete executions. In round $k$, replay restores a shared interaction prefix and context $x$, executes valid action $a$, and lets fixed policy $\pi_{\theta_k}$ complete branch $\tau_k^a$ through actual tool execution and answer submission. Exploration program $\phi$ selects alternatives such as stopping after tool feedback or decreasing or increasing the visual budget. The score and cost cover the chosen action and all subsequent actions. Holding the prefix fixed avoids comparing decisions made with different evidence histories. Training reference answers determine score $s$, whereas deployment uses only queries, acquired evidence, and tool feedback.

\noindent\textbf{Branch comparison.} For action pair $\mathcal A_x=\{a,b\}$, branches must share a verified prefix, start with distinct valid actions, and reach final answers under the same tools and budget. We exclude pairs with two zero scores or identical quality and costs, then select preferred action $a^+$ by Eq.~\eqref{eq:trajectory_objective}. This favors lower cost at equal quality and additional observations when they improve the answer. These comparisons form $\mathcal R_k$ to supervise acquisitions and stopping. We log tool errors separately.

\noindent\textbf{Decision suffixes.} Let $z^+$ and $z^-$ be the preferred and rejected action responses at $x$, including completion tokens. With the same tokenizer and the shared prefix as context, we score each response from the first differing token $j^*$ as
\begin{equation}
g_\theta(z;x)=\sum_{j=j^*}^{|z|}\log\pi_\theta(z_j\mid x,z_{<j}).
\end{equation}
We apply direct preference optimization (DPO)~\cite{rafailov2023dpo} with the frozen parent policy $\pi_{\theta_k}$ as reference. For $\Delta_k(z;x)=g_\theta(z;x)-g_{\theta_k}(z;x)$, the preference loss is
\begin{equation}
\ell_{\mathrm{DPO}}=-\log\sigma\!\left(\beta[\Delta_k(z^+;x)-\Delta_k(z^-;x)]\right),
\label{eq:comparison_loss}
\end{equation}
where $\sigma$ is the logistic sigmoid and $\beta>0$ scales the margin. Shared-prefix terms cancel in this pairwise log-ratio, so suffix scoring retains the full action-response DPO margin. Chosen-suffix cross-entropy reinforces the preferred response, while demonstration cross-entropy rehearses complete tool responses and answers. Let $\ell_{+}=-g_\theta(z^+;x)/(|z^+|-j^*+1)$ and let $\ell_{\mathrm{demo}}$ be the mean token cross-entropy of a demonstration response. We average these losses over sampled examples,
\begin{equation}
\mathcal L_k=\mathbb E_{d\sim\mathcal D_k}
\begin{cases}
\ell_{\mathrm{DPO}}(d)+\ell_{+}(d), & d\in\mathcal R_k,\\
\ell_{\mathrm{demo}}(d), & d\in\mathcal D_{\mathrm{demo}},
\end{cases}
\label{eq:learn}
\end{equation}
Here $\mathcal D_{\mathrm{demo}}$ contains demonstration examples, and $\mathcal D_k$ samples comparisons and demonstrations, including repetitions. DPO sums suffix log-probabilities, whereas each cross-entropy term averages over its supervised tokens. Prompt, tool-feedback, and padding tokens are excluded from the loss. Low-rank adaptation (LoRA) trains adapters with frozen base weights~\cite{r32}.

\subsection{Policy--exploration co-evolution}\label{sec:alternating_updates}
Policy updates change evidence states and observation needs. Fresh rollouts reveal these changes, guiding revisions to the alternatives tested and their exploration budgets. We alternate program revision and policy training, as shown in Figure~\ref{fig:framework}(c), starting from $\pi_{\theta_0}$ initialized with independent teacher and scripted demonstrations, denoted $\Pi_0$ in the results. Each round fixes the policy for program selection, then trains on the selected program's comparisons.

\noindent\textbf{Exploration program revision.} In round $k$, program $\phi_k$ uses branch outcomes, available candidates, and remaining budget to select alternatives or end exploration. The policy executes the selected branches. To adapt these choices, meta-exploration uses a coding agent to revise the exploration program from execution feedback~\cite{r10}, forming a candidate set $\Phi_k$ containing the current program and its revisions. The coding agent's parameters remain fixed. We compare revisions on the same training subset with policy $\pi_{\theta_k}$, action space, and budget $B$ fixed. For each query, Eq.~\eqref{eq:trajectory_objective} selects the best completed trajectory among the original trajectory and explored branches. We aggregate scores with the task weights in Section~\ref{sec:experimental_setup} and average costs across queries. Programs are ranked by this score, then cost, keeping the current program on ties. Fixing the learner isolates changes to exploration.

\noindent\textbf{Policy learning.} The selected program $\phi_{k+1}$ collects comparisons $\mathcal R_k$ through perceptual replay in Section~\ref{sec:observation_comparisons}. Program revision, selection, and comparison collection use separate training subsets with the policy fixed, separating candidate selection from policy supervision. Rounds with fixed exploration collect fresh comparisons with the unchanged program. These comparisons, together with initialization demonstrations, update the policy to $\pi_{\theta_{k+1}}$ via Eq.~\eqref{eq:learn}. Updated states and execution feedback guide subsequent program revision.

After $K$ rounds, $\pi_{\theta_K}$ performs geospatial search with fixed weights, using acquired evidence to return a target set or region answer. Program revision and branch comparison occur only during training. The Supplementary Material details replay, sampling, and the preference objective.

\FloatBarrier
\section{Experiments}
\suppressfloats[t]
\setcounter{topnumber}{2}
\setcounter{bottomnumber}{2}
\setcounter{totalnumber}{4}
We evaluate quality and visual-input savings, the contributions of decision supervision and program revision, and transfer across tasks and policies.

\begin{table*}[t]
\centering
\captionsetup{justification=justified,singlelinecheck=false}
\caption{\textbf{Geospatial search on GeoSearchBench.} The best and second-best scores are \textbf{bold} and \underline{underlined}. Column rankings exclude the tool-based references (structured queries and different access privileges) and the initialization control ($\Pi_0$). Single Query resets evidence before each query. Evidence Reuse replays $N$-query sequences. $P$ counts millions of processed pixels per query. Time is seconds per query, with cached perception unless marked $^{*}$ for online perception including first-use loading. These timings are not a matched speed comparison. Serial reuse Steps are unreported. R1/R2 denote learning rounds, and B denotes billion parameters.}\label{tab:wsvs}
\vspace{-2mm}
\fontsize{8.2}{10}\selectfont
\setlength{\tabcolsep}{1.5pt}\renewcommand{\arraystretch}{1.1}
\setlength{\paperTableWidth}{\linewidth}
\begin{tabularx}{\paperTableWidth}{p{39mm}>{\centering\arraybackslash}p{10.5mm}*{3}{>{\centering\arraybackslash}X}>{\centering\arraybackslash}p{8.5mm}*{6}{>{\centering\arraybackslash}X}}
\toprule\arrayrulecolor{black!70}
\multicolumn{1}{l}{} & \multicolumn{1}{c}{} & \multicolumn{4}{c}{\textbf{Single Query}} & \multicolumn{3}{c}{\textbf{Evidence Reuse} ($N=8$)} & \multicolumn{3}{c}{\textbf{Evidence Reuse} ($N=16$)}\\
\cmidrule(lr){3-6}\cmidrule(lr){7-9}\cmidrule(lr){10-12}
\multicolumn{1}{l}{\multirow{-2}{*}{\textbf{Method}}} & \multicolumn{1}{c!{\paperheaderdivider}}{\multirow{-2}{*}{\textbf{Params.}}} & Score$\uparrow$ & $P\downarrow$ & Time$\downarrow$ & \multicolumn{1}{c!{\paperheaderdivider}}{Steps$\downarrow$} & Score$\uparrow$ & $P\downarrow$ & \multicolumn{1}{c!{\paperheaderdivider}}{Steps$\downarrow$} & Score$\uparrow$ & $P\downarrow$ & Steps$\downarrow$\\\arrayrulecolor{black}\midrule

\compactmethodgroup{Tool-based references}
\methodindent Full-scan Tool+GIS & \textemdash & 60.22 & 2313.81 & 90.61\rlap{$^{*}$} & \textemdash & 60.23 & 288.96 & \textemdash & 60.90 & 144.48 & \textemdash\\
\methodindent Serial retrieval + GIS & \multicolumn{1}{c!{\paperbodydivider{2}}}{\textemdash} & 52.98 & 1745.81 & 70.33\rlap{$^{*}$} & \multicolumn{1}{c!{\paperbodydivider{2}}}{110} & 51.63 & 278.56 & \multicolumn{1}{c!{\paperbodydivider{2}}}{\textemdash} & 52.73 & 141.95 & \textemdash\\
\midrule[0.35pt]
\compactmethodgroup{Predefined acquisition}
\methodindent EvoSearch-Grid & 4B & 37.90 & 215.88 & 66.14\rlap{$^{*}$} & 41 & 35.42 & 128.81 & 41 & 37.97 & 90.37 & 40\\
\methodindent EvoSearch-TopK & \multicolumn{1}{c!{\paperbodydivider{2}}}{4B} & 37.23 & 216.34 & 65.64\rlap{$^{*}$} & \multicolumn{1}{c!{\paperbodydivider{2}}}{41} & 35.31 & 129.13 & \multicolumn{1}{c!{\paperbodydivider{2}}}{41} & 37.57 & 90.69 & 40\\
\midrule[0.35pt]
\compactmethodgroup{Task-adapted VLMs}
\methodindent Molmo2\cite{clark2026molmo2} & 4B & 55.79 & 177.71 & 194.96 & 47 & 54.39 & 119.83 & 46 & 55.90 & 93.59 & 47\\
\methodindent Qwen2.5-VL\cite{r47} & 7B & 54.70 & 156.82 & 149.53 & 40 & 53.25 & 111.81 & 40 & 54.73 & 86.20 & 40\\
\methodindent InternVL3.5-HF\cite{r58} & 8B & 55.28 & 169.83 & 222.23 & 45 & 53.42 & 117.55 & 47 & 55.34 & 89.64 & 44\\
\methodindent LLaVA-OV-1.5\cite{r64} & 8B & 55.75 & 193.24 & 228.28 & 47 & 55.00 & 129.61 & 48 & 56.11 & 98.45 & 47\\
\methodindent Qwen3-VL\cite{r57} & 8B & 55.27 & 185.75 & 226.76 & 47 & 53.53 & 124.72 & 48 & 55.35 & 95.48 & 47\\
\methodindent MiniCPM-V 4.5\cite{r63} & 8B & 53.66 & 131.68 & 159.18 & 34 & 52.10 & 94.70 & 34 & 53.93 & 75.14 & 34\\
\methodindent ZwZ\cite{wei2026zwz} & 8B & 55.01 & 191.07 & 228.33 & 47 & 53.27 & 126.43 & 47 & 55.02 & 97.34 & 46\\
\methodindent Qwen3.5\cite{r56} & \multicolumn{1}{c!{\paperbodydivider{8}}}{9B} & \underline{56.10} & 207.21 & 276.46 & \multicolumn{1}{c!{\paperbodydivider{8}}}{52} & 55.27 & 135.95 & \multicolumn{1}{c!{\paperbodydivider{8}}}{53} & \underline{56.22} & 102.85 & 51\\
\midrule[0.35pt]
\compactmethodgroup{Ours (Qwen3.5-4B)}
\methodindent EvoSearch-$\Pi_0$ & 4B & 56.10 & 208.08 & 314.42 & 52 & 55.27 & 136.62 & 53 & 56.22 & 103.15 & 51\\
\methodindent EvoSearch-R1 & 4B & 55.34 & 160.02 & 225.39 & 38 & 54.50 & 111.58 & 39 & 55.10 & 86.17 & 37\\
\variantindent Fixed exploration (S2) & 4B & 55.84 & 149.30 & 333.48 & 55 & \underline{55.88} & 103.18 & 55 & 55.75 & 80.52 & 54\\
\rowcolor{ourrow}\methodindent \textbf{EvoSearch (R2)} & \multicolumn{1}{c!{\paperbodydivider{4}}}{4B} & \textbf{57.11} & 113.31 & 150.20 & \multicolumn{1}{c!{\paperbodydivider{4}}}{27} & \textbf{55.91} & 85.49 & \multicolumn{1}{c!{\paperbodydivider{4}}}{27} & \textbf{56.94} & 66.67 & 26\\
\bottomrule
\end{tabularx}
\vspace{-3mm}
\end{table*}

\subsection{Experimental setup}\label{sec:experimental_setup}

\noindent\textbf{Benchmarks.} GeoSearchBench, summarized in Figure~\ref{fig:showcase}(c), evaluates relational and region search with disjoint training, validation, and test regions. Decision-supervision comparisons use 50 validation queries from eight regions, while main results use 368 test queries from six regions. The Supplementary Material details imagery, query construction, spatial reference answers, and scoring. Additional benchmarks are EarthVQA~\cite{r22} with 26,704 selected validation questions, XLRS-Bench-lite~\cite{r24} (XLRS-lite) with 3,080 questions across 13 tasks, and V*Bench~\cite{r1} with 191 questions requiring fine visual detail.

\noindent\textbf{Evaluation metrics.} Relational search uses set F1 to measure agreement between returned and reference targets. Region search requires exactly the reference set of regions. GeoSearchBench averages these two task scores equally. VQA benchmarks report accuracy, averaging equally across the 13 XLRS-lite tasks. Quality scores are percentages. Visual input counts pixels processed by visual encoders per query, including repeats, and is reported in millions of pixels. Runtime is reported in seconds per query. Tool-call counts include submission. GeoSearchBench training prefers fewer high-resolution acquisitions, then fewer steps, among equal-quality branches. Encoded pixels measure the resulting visual workload. Unanswered queries are included when computing all reported averages. Single Query clears evidence, dialogue, and visual features before each query. Evidence Reuse replays saved trajectories over 8 or 16 queries. The Supplementary Material specifies tool access, replay, and cost accounting.

\noindent\textbf{Baselines.} In Table~\ref{tab:wsvs}, Full-scan Tool+GIS examines all imagery and Serial retrieval + GIS follows fixed low-resolution scene rankings. Both receive structured queries. Grid and TopK use predefined spatial and ranked scene orders. Task-adapted VLMs share tools, demonstrations, and initialization budgets with our initial policy. Table~\ref{tab:publicmain} compares general and remote-sensing VLMs, visual-search systems, and token-pruning methods on image-based VQA.

\noindent\textbf{Implementation details.} Qwen3.5-4B is initialized with fresh LoRA using independent teacher and scripted demonstrations. GeoSearchBench uses two learning rounds with screened tool interactions and answers. Fixed exploration collects fresh experience with the current policy and unchanged program, matching training updates. Full is the selected learned policy.

\begin{table*}[t]
\centering\begingroup
\captionsetup{justification=justified,singlelinecheck=false}
\caption{\textbf{Policy learning on image-based VQA benchmarks.} The best score is highlighted in \textbf{bold}, and the second is \underline{underlined}. Rankings exclude tool-initialization controls ($\Pi_0$). $P$ counts millions of processed pixels per query, and Time is seconds per query. Full denotes the selected policy. Avg. weights benchmarks equally. $^{\dagger}$~Cross-domain evaluation of released remote-sensing checkpoints on V*Bench, without additional task-specific fine-tuning.}\label{tab:publicmain}
\vspace{-2mm}
\fontsize{8.2}{10}\selectfont
\setlength{\tabcolsep}{1.5pt}\renewcommand{\arraystretch}{1.0}
\setlength{\paperTableWidth}{\linewidth}
\begin{tabularx}{\paperTableWidth}{p{39mm}c*{3}{>{\centering\arraybackslash}X}*{3}{>{\centering\arraybackslash}X}*{3}{>{\centering\arraybackslash}X}*{2}{>{\centering\arraybackslash}X}}
\toprule\arrayrulecolor{black!70}
\multicolumn{1}{l}{\paperheaderbg{2}} & \multicolumn{1}{c}{} & \multicolumn{3}{c}{\textbf{EarthVQA}} & \multicolumn{3}{c}{\textbf{XLRS-lite}} & \multicolumn{3}{c}{\textbf{V*Bench}} & \multicolumn{2}{c}{\textbf{Avg.}}\\
\cmidrule(lr){3-5}\cmidrule(lr){6-8}\cmidrule(lr){9-11}\cmidrule(lr){12-13}
\multicolumn{1}{l}{\multirow{-2}{*}{\textbf{Method}}} & \multicolumn{1}{c!{\paperheaderdivider}}{\multirow{-2}{*}{\textbf{Params.}}} & Score$\uparrow$ & $P\downarrow$ & \multicolumn{1}{c!{\paperheaderdivider}}{Time$\downarrow$} & Score$\uparrow$ & $P\downarrow$ & \multicolumn{1}{c!{\paperheaderdivider}}{Time$\downarrow$} & Score$\uparrow$ & $P\downarrow$ & \multicolumn{1}{c!{\paperheaderdivider}}{Time$\downarrow$} & Score$\uparrow$ & $P\downarrow$\\\arrayrulecolor{black}\midrule
\compactmethodgroup{General and remote-sensing VLMs}
\methodindent EarthDial\cite{earthdial2025} & 4B & 77.16 & 1.00 & 0.46 & 41.77 & 1.09 & 2.94 & 37.70\rlap{$^{\dagger}$} & 1.33 & 0.64 & 52.21 & 1.14\\
\methodindent Molmo2\cite{clark2026molmo2} & 4B & 77.54 & 2.43 & 0.61 & 41.50 & 2.70 & 1.69 & 83.25 & 3.54 & 0.73 & 67.43 & 2.89\\
\methodindent Qwen2.5-VL\cite{r47} & 7B & 75.63 & 1.07 & 0.56 & 48.45 & 12.90 & 15.69 & 83.77 & 14.42 & 5.90 & 69.28 & 9.46\\
\methodindent LLaVA-OV (Direct)\cite{li2024llavaonevision} & 7B & \underline{78.51} & 1.47 & 1.53 & 43.01 & 5.18 & 3.68 & 75.39 & 1.90 & 1.61 & 65.64 & 2.85\\
\methodindent GeoLLaVA-8K\cite{r21} & 7B & 72.22 & 1.81 & 0.35 & 50.61 & 52.24 & 8.07 & 37.17\rlap{$^{\dagger}$} & 4.09 & 0.63 & 53.33 & 19.38\\
\methodindent Qwen3-VL & 8B & 77.99 & 1.05 & 1.01 & 46.37 & 12.87 & 13.20 & 84.82 & 3.49 & 5.55 & 69.73 & 5.80\\
\methodindent Qwen3.5\cite{r56} & 4B & 78.27 & 1.05 & 0.49 & 45.45 & 17.02 & 9.92 & 77.49 & 3.49 & 1.11 & 67.07 & 7.19\\
\methodindent InternVL3.5-HF\cite{r58} & 8B & 78.35 & 2.01 & 0.68 & \underline{51.54} & 1.95 & 1.75 & 72.25 & 1.57 & 0.46 & 67.38 & 1.84\\
\methodindent ZwZ\cite{wei2026zwz} & 8B & 75.67 & 1.05 & 0.87 & 48.91 & 17.03 & 13.65 & 90.05 & 3.49 & 3.23 & 71.54 & 7.19\\
\methodindent MiniCPM-V 4.5\cite{r63} & \multicolumn{1}{c!{\paperbodydivider{10}}}{8B} & 76.76 & 1.40 & \multicolumn{1}{c!{\paperbodydivider{10}}}{0.55} & 44.20 & 2.02 & \multicolumn{1}{c!{\paperbodydivider{10}}}{4.37} & 70.68 & 1.85 & \multicolumn{1}{c!{\paperbodydivider{10}}}{1.12} & 63.88 & 1.76\\
\midrule[0.35pt]
\compactmethodgroup{Visual search and token pruning}
\methodindent ZoomEarth\cite{r3} & 3B & 71.27 & 1.07 & 0.87 & 42.43 & 2.16 & 16.65 & 60.73\rlap{$^{\dagger}$} & 1.61 & 14.15 & 58.14 & 1.61\\
\methodindent HiDe\cite{hide2025} & 7B & 72.57 & 2.47 & 2.17 & 51.27 & 33.48 & 16.16 & \underline{91.10} & 7.31 & 5.46 & \underline{71.65} & 14.42\\
\methodindent Pixel Reasoner\cite{r15} & 7B & 75.64 & 1.07 & 1.06 & 46.23 & 7.05 & 20.21 & 84.29 & 5.27 & 7.62 & 68.72 & 4.46\\
\methodindent VisionZip (Qwen2.5-VL)\cite{yang2025visionzip} & 7B & 75.48 & 1.07 & 0.93 & 49.26 & 12.90 & 12.40 & 67.54 & 1.56 & 0.61 & 64.09 & 5.18\\
\methodindent Coarse-to-Fine\cite{luo2025coarsefine} & \multicolumn{1}{c!{\paperbodydivider{5}}}{7B} & 77.32 & 2.94 & \multicolumn{1}{c!{\paperbodydivider{5}}}{0.49} & 40.30 & 3.02 & \multicolumn{1}{c!{\paperbodydivider{5}}}{4.09} & 50.79\rlap{$^{\dagger}$} & 2.40 & \multicolumn{1}{c!{\paperbodydivider{5}}}{0.46} & 56.14 & 2.79\\
\midrule[0.35pt]
\compactmethodgroup{Ours}
\methodindent Qwen3.5 + tools ($\Pi_0$) & 4B & 78.27 & 1.11 & 4.23 & 45.95 & 20.72 & 17.14 & 83.25 & 5.00 & 33.38 & 69.16 & 8.94\\
\methodindent Qwen3.5 + EvoSearch (Full) & 4B & \textbf{78.60} & 0.94 & 3.56 & 45.06 & 5.18 & 6.70 & 84.29 & 4.68 & 34.19 & 69.32 & 3.60\\
\methodindent HiDe + tools ($\Pi_0$) & 7B & 72.57 & 2.49 & 10.21 & 51.27 & 32.80 & 19.52 & 93.19 & 7.32 & 7.17 & 72.34 & 14.20 \\
\rowcolor{ourrow}\methodindent HiDe + EvoSearch (Full) & \multicolumn{1}{c!{\paperbodydivider[3.8pt]{4}}}{7B} & 72.90 & 2.33 & \multicolumn{1}{c!{\paperbodydivider[3.8pt]{4}}}{8.96} & \textbf{51.62} & 31.45 & \multicolumn{1}{c!{\paperbodydivider[3.8pt]{4}}}{18.36} & \textbf{94.24} & 6.91 & \multicolumn{1}{c!{\paperbodydivider[3.8pt]{4}}}{6.38} & \textbf{72.92} & 13.56 \\
\bottomrule
\end{tabularx}
\vspace{-2mm}
\endgroup\end{table*}

\subsection{Main results}\label{sec:main_results}

\noindent\textbf{Search performance.} We first test whether learning preserves useful evidence while reducing visual input. After two rounds, EvoSearch processes 45.54\% less visual input per independent query and raises the answer score by 1.8\% relative to initialization, as Table~\ref{tab:wsvs} and Figure~\ref{fig:showcase}(d) show. Both policies share tools and observation sizes, so savings reflect learned acquisition under the same interface. EvoSearch also achieves better answers with less visual input than every task-adapted VLM. Grid and TopK process roughly as much imagery as initialization, yet score over 32\% lower. Observation choices determine how effectively the input budget supplies evidence. EvoSearch also surpasses serial retrieval in both quality and visual efficiency. EvoSearch uses about one twentieth of the input required by exhaustive coverage, with answer scores of 57.11\% and 60.22\%, respectively.

\begin{figure}[t]
\centering
\includegraphics[width=0.99\linewidth]{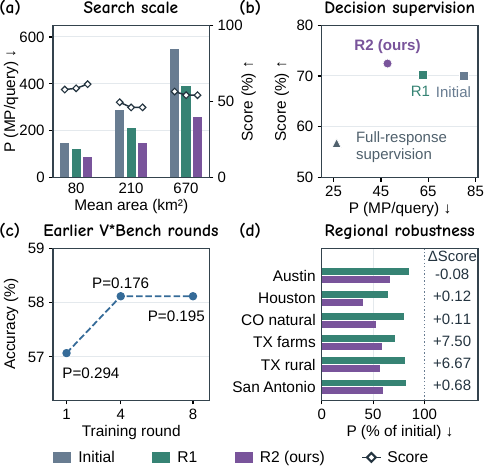}
\caption{\textbf{Policy learning analysis.} (a) Search scale, with mean query areas rounded to ten square kilometers. (b) Decision supervision on 50 validation queries. (c) Accuracy by learning round in the earlier V*Bench setup. Labels give visual input in millions of pixels per query. (d) Regional input relative to initialization. Score differences subtract R1 from R2 on the percentage scale. CO/TX denote Colorado/Texas.}\label{fig:experience_analysis_current}
\vspace{-2mm}
\end{figure}

\noindent\textbf{Search scale.} To test whether savings persist over wider areas, Figure~\ref{fig:experience_analysis_current}(a) groups queries by search area. Visual input falls below initialization in all three groups, with larger absolute savings over wider areas. Relative to round one, round two matches or improves quality with less visual input at every scale. Compared with initialization, quality improves in the smallest-area group and remains lower in the larger groups. Input savings persist across scales, while quality recovery varies.

\noindent\textbf{Regional robustness.}\label{sec:cross_region} Regional comparisons check whether gains concentrate in a few locations. Between rounds, visual input decreases in all six regions and answer quality improves in five, as shown in Figure~\ref{fig:experience_analysis_current}(d). Texas farms and Texas rural gain most in quality, while Austin declines slightly. Across these geographic settings, observation learning consistently reduces visual input.

\noindent\textbf{Evidence reuse.} To separate learning from observation reuse, we compare initial and learned policies under identical reuse rules. Table~\ref{tab:wsvs} shows over 35\% less visual input and slightly better answers with learning at both sequence lengths. The learned policy thus reduces the visual input needed even when earlier observations can be reused.

\noindent\textbf{Changes across learning rounds.} We examine whether later updates refine the initial quality--cost trade-off. Round two recovers the first round's quality loss and exceeds initialization for independent queries and both reuse lengths, while reducing input further. Later updates thus recover quality while retaining input savings.

\begin{figure*}[t]\centering
\includegraphics[width=\textwidth]{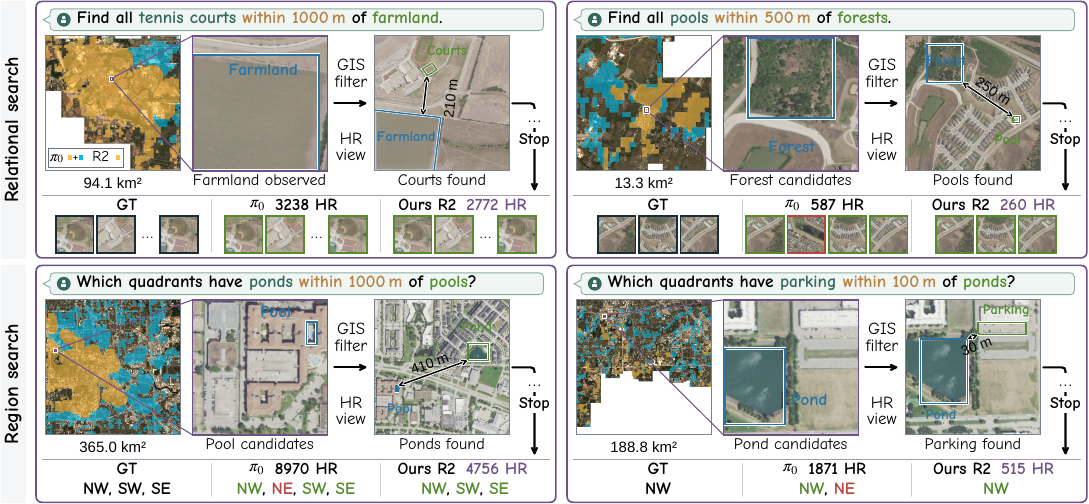}\par
\vspace{-1mm}
\caption{\textbf{Qualitative geospatial search examples.} Relational search in San Antonio and Austin is shown above region search in Houston. GT denotes ground truth. Gold marks R2 coverage, and cyan marks additional $\Pi_0$ coverage. HR counts unique observed high-resolution tiles. Areas below the maps refer to the query regions. Object outlines and approximate distances provide geographic context.}\label{fig:cases_current}
\vspace{-3mm}
\end{figure*}

\noindent\textbf{Qualitative results.} Figure~\ref{fig:cases_current} examines which evidence selective observation retains.\label{sec:qualitative_results} Farmland and forests locate where nearby tennis courts and pools must be checked. The learned policy recovers the illustrated matching targets with fewer high-resolution observations. For ponds near pools, the learned policy examines 4,214 fewer high-resolution tiles and excludes the northeast region incorrectly returned by initialization, recovering exactly the three reference regions. For parking near ponds, it likewise removes the incorrect northeast region and retains northwest. Selective observation retains evidence for finding matching targets and identifying the correct query regions.

\subsection{Ablation studies}\label{sec:ablation_studies}\label{sec:alternating_updates_evaluation}
We examine decision supervision in Figure~\ref{fig:experience_analysis_current}(b), program revision in Table~\ref{tab:wsvs}, and candidate selection in Table~\ref{tab:exploration_supervision}.

\noindent\textbf{Decision supervision.} We test whether decision supervision preserves useful evidence more effectively than full-response supervision. Both start from the first-round policy with matched replay data and training budgets. In Figure~\ref{fig:experience_analysis_current}(b), full-response supervision gives the largest visual-input reduction but the poorest answers. Decision supervision improves both quality and visual efficiency over the first round. With matched replay data, this comparison supports focusing supervision on the decisions that determine which evidence to acquire. The Supplementary Material specifies the compared losses and supervised tokens.

\noindent\textbf{Exploration program revision.} To distinguish program revision from further training, both variants collect fresh current-policy experience and receive equal training updates. Only one revises its exploration program. In Table~\ref{tab:wsvs}, revision reduces visual input by 24.1\% and raises the answer score by 2.3\% relative to fixed exploration for independent queries. The matched updates support adapting experience collection to the current policy. The Supplementary Material reports experience collection and policy training costs separately. With reusable evidence, revision saves about 17\% of visual input and roughly halves calls at both lengths. Quality remains similar over eight-query sequences and improves over sixteen-query sequences. \par\noindent\textbf{Program selection.} We test whether program changes produce better completed trajectories before policy training. Table~\ref{tab:exploration_supervision} compares programs on 24 shared training queries with the policy and budget fixed. The first revision yields better answers with less visual input and is adopted. In round two, selection retains the current program for its better quality and lower input. Trajectory comparisons thus determine when to revise the program.

\begin{table}[t]
\centering\small
\captionsetup{justification=justified,singlelinecheck=false}
\caption{\textbf{Program selection during meta-exploration.} Programs share 24 training queries, policy, and budget per round. $P$ measures visual input in millions of pixels per query.}\label{tab:exploration_supervision}
\vspace{-2mm}
\setlength{\tabcolsep}{5pt}\renewcommand{\arraystretch}{1.12}
\begin{tabularx}{\linewidth}{cXrrc}
\toprule
\textbf{Round} & \textbf{Program} & \textbf{Score}$\uparrow$ & $\boldsymbol{P}\downarrow$ & \textbf{Selected}\\
\midrule
1 & Current & 70.94 & 91.93 & ---\\
\rowcolor{ourrow}1 & Revision & 74.06 & 75.24 & Yes\\
\midrule[0.35pt]
\rowcolor{ourrow}2 & Current & 70.94 & 61.28 & Yes\\
2 & Revision & 70.11 & 68.30 & ---\\
\bottomrule
\end{tabularx}
\vspace{-2mm}
\end{table}

\begin{table}[t]
\centering\small
\captionsetup{justification=justified,singlelinecheck=false}
\caption{\textbf{Policy learning with Pixel Reasoner-7B on V*Bench.} S2/R2 use fixed/revised exploration with matched initialization and updates. $P$ counts millions of pixels per query.}\label{tab:pixel_transfer_main}
\vspace{-2mm}
\setlength{\tabcolsep}{4pt}\renewcommand{\arraystretch}{1.09}
\setlength{\paperTableWidth}{\linewidth}
\begin{tabularx}{\paperTableWidth}{p{0.48\paperTableWidth}*{2}{>{\centering\arraybackslash}X}}
\toprule
\paperheaderbg{1}\textbf{Variant} & \textbf{Score}$\uparrow$ & $\boldsymbol{P}\downarrow$\\
\midrule
Released model~\cite{r15} & 84.29 & 5.27\\
\midrule[0.35pt]
Adapted initial policy ($\Pi_0$) & 81.68 & 6.73\\
\quad + Fixed exploration (S2) & 84.82 & 3.41\\
\rowcolor{ourrow}\quad \textbf{+ EvoSearch (R2)} & 85.34 & 3.41\\
\bottomrule
\end{tabularx}
\vspace{-2mm}
\end{table}

\begin{figure}[t]
\centering
\includegraphics[width=0.97\linewidth]{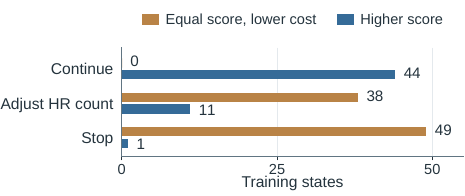}\par
\vspace{-2mm}
\caption{\textbf{Decision preferences.} Counts from 143 unique R1/R2 states. Cost preferences require equal, nonzero scores. Higher-score preferences may retain the original action.}\label{fig:training_preferences_main}
\vspace{-4mm}
\end{figure}

\noindent\textbf{Decision preferences.} We test whether supervision rewards useful observation as well as stopping. In Figure~\ref{fig:training_preferences_main}, stopping accounts for 56\% of equal-score preferences, while continuing accounts for 79\% of higher-score preferences. Preferences favor both efficient stopping and useful observation.

\noindent\textbf{Evolution rounds.} We test whether more rounds improve learning. In the earlier V*Bench setup in Figure~\ref{fig:experience_analysis_current}(c), round four improves accuracy over round one with about 40\% less visual input. Round eight uses about 11\% more input without improving accuracy. Round four therefore gives the best quality--input balance among the three shown. This earlier setup is distinct from Table~\ref{tab:publicmain}.

\subsection{Transfer evaluation}\label{sec:observation_analysis}
\noindent\textbf{Task transfer.} Table~\ref{tab:publicmain} tests transfer beyond geospatial search against matched tool initializations. Visual input decreases in all six comparisons, and accuracy improves in five. HiDe gains accuracy on every benchmark with 4.1--6.4\% less input, extending observation learning to an existing visual-search system. HiDe has the highest average accuracy, while Qwen3.5 uses less imagery. Qwen3.5 achieves its largest input reduction on XLRS-lite, saving 75\% with mean task accuracy of 45.06\%, compared with 45.95\% initially. On V*Bench, reduced input accompanies a 0.81-second runtime increase per query. Input and latency capture different costs. The Supplementary Material details task-level accuracy and timing.

\noindent\textbf{Policy transfer.} Table~\ref{tab:pixel_transfer_main} separates tool adaptation from learning for Pixel Reasoner-7B. From the adapted initialization, two learning rounds add seven correct answers while nearly halving input. Both fixed and revised exploration surpass the released model with less imagery, improving beyond the quality lost during tool adaptation. With updates matched, revision adds one correct answer at the same visual input as fixed exploration. The framework thus improves another backbone after task-specific initialization.

\section{Conclusion}

We introduced EvoSearch, a self-evolving active perception framework for scalable geospatial search. By combining executable tool actions with fixed-size multi-scale observations, we formulate active perception as quality-preserving trajectory compression. Perceptual replay provides decision supervision through trajectory comparisons, while meta-exploration adapts experience generation to execution feedback, enabling policy--exploration co-evolution. Experiments demonstrate improved quality--cost trade-offs, with ablations supporting both mechanisms. These results suggest that adaptive observation decisions can reduce redundant visual processing in large-scale geospatial search.

\FloatBarrier
\clearpage
\section*{LLM Usage Statement}
OpenAI Codex assisted with manuscript editing, coding, and LaTeX formatting. Model use within the proposed method is described in the paper. The authors take responsibility for the final manuscript.\par

{\small\widowpenalty=1000\clubpenalty=1000
}
\clearpage
\end{document}